# Evaluating Subword Tokenization Techniques for Bengali: A Benchmark Study with BengaliBPE

**Firoj Ahmmed Patwary**
Department of Mathematics & Computer Science
Freie Universität Berlin
14195 Berlin, Germany
`firoj.patwary@fu-berlin.de`

**Abdullah Al Noman**
BIBA-Bremer Institut für Produktion und Logistik GmbH
University of Bremen
28359 Bremen, Germany
`noman@uni-bremen.de`

November 7, 2025

## ABSTRACT

Tokenization is an important first step in Natural Language Processing (NLP) pipelines because it decides how models learn and represent linguistic information. But current subword tokenizers like SentencePiece or HuggingFace BPE are mostly made for Latin or multilingual corpora and don't work well on languages with a lot of morphology, like Bengali. To fill this gap, we present BengaliBPE, a Byte Pair Encoding (BPE) tokenizer made just for the Bengali script. BengaliBPE uses Unicode normalization, grapheme-level initialization, and morphology-aware merge rules to keep the language consistent and the subwords intact. We use a large-scale Bengali news classification dataset to compare BengaliBPE to three other systems: Whitespace, SentencePiece BPE, and HuggingFace BPE. The evaluation looks at how fine the tokenization is, how fast the encoding is, and how accurate the downstream classification is. All of the methods work well overall (macro-F1 $\geq$ 0.88), but BengaliBPE gives the most detailed segmentation and the best morphological interpretability, even though it costs a little more to run. These results show how important language-aware tokenization is for scripts that are morphologically complex. They also show that BengaliBPE is a good starting point for future Bengali NLP systems, such as large-scale pretraining of contextual models.

## 1 Introduction

Tokenization is a key part of modern Natural Language Processing (NLP) because it connects raw text with units that machines can understand [19]. The quality of tokenization has a direct effect on how well downstream models capture linguistic structure and meaning. Subword-based tokenization methods like Byte Pair Encoding (BPE) [7] and WordPiece [20] work well for languages with a lot of morphology, like German or Turkish. However, they don't work as well for Indic scripts, like Bengali. The main reasons for this are differences in spelling, the difficulty of making compound words, and the fact that there aren't any standard ways to preprocess Bengali text.

Bengali is the seventh most spoken language in the world, with more than 234 million native speakers [6, 18]. However, it is still not well represented in the creation of basic NLP tools. Tokenizers like SentencePiece [11] and HuggingFace's BPE [8] are usually trained on multilingual corpora that work best with Latin or mixed scripts. Because of this, these tokenizers often don't split Bengali words correctly, combine subwords that look similar but have different meanings, or deal with the natural variability of Unicode compositions in Bengali text. These kinds of inconsistencies spread through the NLP pipeline, which hurts the quality of embeddings and the performance of downstream tasks like classification, translation, or generation.

To solve these problems, we present BengaliBPE, a Byte Pair Encoding tokenizer that only works with Bengali text. BengaliBPE uses Unicode normalization, script-aware preprocessing, and a merge strategy that works better with the way Bengali words are formed. We use a large-scale Bengali news classification dataset to systematically compare it to three well-known baselines Whitespace, SentencePiece BPE, and HuggingFace BPE to see how well it works. When evaluating a tokenizer, we look at more than just how well it segments and how fast it can process data. We also look at how it affects the performance of the models that come after it. Our experiments indicate that, although surface-level classification accuracy is consistent across all methods, BengaliBPE offers more linguistically significant segmentation and enhanced potential for neural language model integration.

The main contributions of this study are as follows:

- We propose BengaliBPE, the first open-source, Bengali-specific Byte Pair Encoding tokenizer that integrates linguistic normalization and morphology-aware merging.
- We provide a comprehensive benchmark of four tokenization techniques for Bengali text using both quantitative (F1, accuracy, speed) and qualitative (morphological consistency) measures.
- We release the BengaliBPE package publicly on PyPI to support open research and reproducibility in Bengali NLP.[1]

## 2 Related Work

Since tokenization controls the division of unprocessed text into distinct units that can be fed into models, it continues to be a fundamental part of Natural Language Processing (NLP) pipelines. Typically, early systems relied on punctuation or whitespace delimiters and employed word-level tokenization. For high-resource languages with less morphological complexity and comparatively fixed orthography, this method performs admirably. However, word-level segmentation frequently results in very large vocabularies and high out-of-vocabulary (OOV) rates for morphologically rich languages, like Bengali, which has compounding, inflection, and agglutination, restricting generalization.

Subword tokenization techniques arose to solve the vocabulary sparsity and OOV issues, and they eventually took center stage. The use of Byte Pair Encoding (BPE) for neural machine translation by [17] is among the first significant works in NLP that used subword tokenization. [10] introduces a subword sampling technique that generates multiple segmentations to prevent overfitting in neural translation. In order to facilitate open-vocabulary modeling and effective handling of rare words, BPE iteratively combines the most common symbol pairs in the training corpus to create a compact vocabulary of subword units. Since then, the algorithm has been widely used in large language model tokenizers such as RoBERTa [12] and GPT [14].

Further tokenization techniques built upon BPE by introducing various training objectives and probabilistic models. In order to pre-train Google's BERT [4], the WordPiece algorithm [16] divides words into subwords using a probabilistic model that is tailored for tasks that come after. Similarly, [11] SentencePiece framework is language-agnostic, appropriate for scripts and low-resource environments, and supports both BPE and unigram language-model(LM) tokenization. According to recent research by [3], a unigram LM tokenization approach can more closely match morphology, particularly for complex scripts, and BPE may not be the best option for language-model pretraining. Recently, [21] provided a more formal analysis of BPE's optimization properties, demonstrating that its greedy merging is a sub-optimal approximation to a combinatorial optimum.

The particular difficulties of Indic languages have spurred specialized tokenization and resource work in addition to algorithmic advancements. For Indian languages, the Indic NLP Library [9] offers script conversion, normalization, and tokenization support; however, it employs general-purpose techniques and does not customize merges for individual languages. Large-scale monolingual corpora and evaluation benchmarks for many Indian languages were introduced by the IndicNLP Suite [9]. Models like IndicBERT [5], a multilingual ALBERT variant trained on 12 Indian languages, including Bengali, were also released. Multilingual tokenizers frequently under-segment or incorrectly split morphologically coherent units in Indic scripts, leading to higher fertility (more subwords per word) or less interpretable splits, according to evaluations of IndicBERT and related tokenizers [5]. [2] proposes a monolingual BERT model for Bengali with extensive benchmarking across various NLP tasks.

Despite these efforts, tokenizers for Indic languages still use the same tokenization heuristics across scripts and are trained on aggregated multilingual corpora, making them language-agnostic. Because of this, they usually miss morphological or orthographic features unique to the language, like Bengali conjunct consonants, dependent vowel signs, and rich affixation. For instance, tokenizers trained cross-lingually produce longer token sequences and less morphological alignment in scripts such as Bengali and Assamese, according to small studies [13].

Our suggested tokenizer, BengaliBPE, on the other hand, is a linguistically informed, monolingual subword tokenizer created especially for the Bengali script. By combining merge-constraint heuristics specific to Bengali morphology, grapheme-based initialization, and Unicode normalization, it aligns segmentation units with significant linguistic boundaries. BengaliBPE's targeted design enables it to bridge the crucial gap in script-specific tokenization for Indic languages and enhance current multilingual tokenizers.

---

[1] https://pypi.org/project/bengali-bpe/

# 3 Methodology

## 3.1 Overview

The suggested BengaliBPE framework builds on the traditional Byte Pair Encoding algorithm by adding rules for Bengali-specific preprocessing, character normalization, and subword merging. The method seeks to generate linguistically coherent subword units that more accurately reflect the morphology of Bengali, encompassing inflectional suffixes, compound words, and reduplicated forms. Figure 1 (conceptual) shows the whole process of BengaliBPE, from normalizing text and building a vocabulary to encoding and decoding.

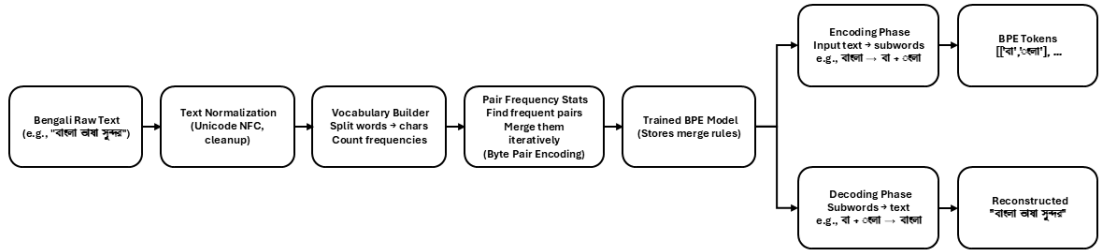

Figure 1: Conceptual workflow of BengaliBPE, illustrating the complete process from raw Bengali text normalization and vocabulary construction to BPE model training, encoding into subword units, and final decoding to reconstruct the original text.

## 3.2 Preprocessing for Bengali Text

Before the tokenization process, the raw Bengali corpus undergoes a preprocessing stage tailored to the unique properties of the Bengali script. This includes:

- **Unicode normalization:** All text is standardized using NFKC normalization to ensure consistent representation of visually identical characters that may differ in byte form.

- **Character filtering:** Non-Bengali characters, numerals, and extraneous symbols are removed using regular expressions that retain only Bengali letters (Unicode range: U+0980U+09FF), punctuation, and whitespace.

- **Whitespace normalization:** Consecutive spaces and irregular spacing patterns are reduced to a single space.

- **Optional lower-level cleaning:** URLs, emojis, and HTML tags are removed for corpus consistency.

This preprocessing ensures that the tokenizer learns from a uniform and linguistically valid representation of the Bengali language.

## 3.3 BengaliBPE Algorithm

BengaliBPE is based on the standard BPE principle proposed by [17], which iteratively merges the most frequent pairs of symbols in a corpus to build a vocabulary of subwords. However, unlike traditional implementations, BengaliBPE introduces the following modifications:

- **Script-aware initialization:** Instead of starting from raw characters, BengaliBPE initializes the symbol vocabulary using normalized Bengali graphemes. This step prevents the accidental splitting of consonant clusters or dependent vowel signs, which are frequent in Bengali.

- **Language-informed merge strategy:** Merge operations are restricted to linguistically plausible boundaries (e.g., between roots and suffixes), reducing nonsensical token merges commonly found in generic BPE models.

- **Unicode-preserving encoding:** During encoding, BengaliBPE ensures that diacritic marks (kar and phala forms) are not detached from their base consonants, preserving visual and phonetic integrity.

- **Python-based implementation:** The algorithm is implemented in pure Python, allowing easy integration with downstream workflows, debugging, and educational use. Despite being slower than optimized C++ implementations, it provides transparency and full control over the tokenization process.

**Table 1: Example of Subword Segmentation Across Tokenizers**

| Tokenizer | Tokenized Output |
|-----------|------------------|
| Original Sentence | বাংলা ভাষাভাষীরা গর্বিত। |
| Whitespace | [বাংলা, ভাষাভাষীরা, গর্বিত।] |
| SentencePiece BPE | [বাংলা, ভাষা, ভাষী, রা, গর্বিত, ।] |
| HuggingFace BPE | [বাংলা, ভাষাভা, ষীরা, গর, বিত, ।] |
| BengaliBPE (Proposed) | [বাংলা, ভাষা, ভাষী, রা, গর, বিত, ।] |

Table 1: Example of subword segmentation for a Bengali sentence using different tokenizers. BengaliBPE produces linguistically coherent subwords that align with morphological boundaries, while generic BPE methods often produce arbitrary splits.

### 3.4 Training and Encoding Procedure

First, the training corpus is normalized and broken down into individual characters or graphemes. We count how many times each pair of adjacent symbols appears, and then we merge the most common pair over and over until we reach a certain vocabulary size or number of merges. The merge rules that come out of this are saved in a merge table and used to encode new text. Each sentence is broken down into subwords during encoding based on these learned rules. This makes sure that unknown words can still be represented as combinations of existing subword units.

We trained BengaliBPE with a target vocabulary of 24,000 subwords for benchmarking. This is similar to the SentencePiece and HuggingFace BPE settings. We tested the model on a large set of Bengali news articles and compared its outputs to those of baseline tokenizers that used the same preprocessing and classification pipelines.

### 3.5 Comparison Baselines

To contextualize the performance of BengaliBPE, three additional tokenization approaches were included in the benchmark:

- **Whitespace tokenizer:** Simple word-level segmentation based on space delimiters.

- **SentencePiece BPE:** A data-driven unsupervised BPE tokenizer with a unigram fallback mechanism, widely used in multilingual NLP.

- **HuggingFace BPE:** The subword tokenizer used in models such as RoBERTa and GPT, optimized for performance and multilingual integration.

All tokenizers were trained on the same preprocessed corpus and evaluated using identical hyperparameters to ensure a fair comparison.

## 4 Experimental Setup

### 4.1 Dataset

To assess the efficacy of the proposed BengaliBPE tokenizer, experiments were performed utilizing a substantial open-source Bengali news classification[1] dataset named 'Potrika'. The dataset has news stories from a number of well-known Bengali news sites. These stories cover a wide range of topics, including politics, sports, economy, entertainment, education, national, international, science and technology. After the first cleaning and preprocessing, the final dataset had about 327126 labeled news articles.

Each record in the dataset has two parts: the text itself (text) and the category label that goes with it (label). To make sure that the Unicode representation was consistent and that non-Bengali artifacts were removed, the Bengali preprocessing pipeline described in Section 3.2 was used to normalize the text data. We split the dataset into three parts: training, validation, and testing. The training set had 70% of the data, the validation set had 10%, and the testing set had 20%. This kept the class distributions across categories balanced.

This dataset serves as a realistic benchmark for Bengali tokenization research, as it embodies naturally occurring linguistic phenomena, including compound words, affixation, and orthographic variations prevalent in contemporary Bengali media.

**Table 2: Bengali News Dataset (Potrika) Statistics**

| Category | Number of Articles | Percentage (%) |
|---|---|---|
| Politics | 40,179 | 12.21% |
| Sports | 41,000 | 12.46% |
| Economy | 40,848 | 12.41% |
| Entertainment | 40,772 | 12.39% |
| Education | 40,916 | 12.43% |
| National | 41,000 | 12.46% |
| International | 41,000 | 12.46% |
| Science & Technology | 43,395 | 13.18% |

Table 2: Statistics of the Bengali News Classification Dataset (Potrika). The dataset contains eight balanced categories and was divided into 70% training, 10% validation, and 20% testing splits, ensuring equal representation across all classes.

### 4.2 Tokenizers Evaluation

Four tokenization approaches were evaluated in this study:

- **Whitespace Tokenizer:** A simple segmentation strategy using spaces as delimiters. This baseline represents word-level tokenization and serves as a performance ceiling for coarse-grained text classification tasks.

- **SentencePiece BPE:** A widely used subword tokenizer that employs unsupervised learning of merge rules from corpus statistics. It is language-agnostic and implemented in C++, offering efficient processing.

- **HuggingFace BPE:** The tokenizer framework used in Transformer-based models such as GPT and RoBERTa. It applies a similar frequency-based merge process to SentencePiece but is optimized for integration with modern deep learning toolkits.

- **BengaliBPE (Proposed):** Our language-specific implementation that incorporates Bengali Unicode normalization, grapheme-based initialization, and linguistically informed merge rules to preserve morphemic boundaries. BengaliBPE was trained on the same corpus as the baselines, with a target vocabulary of 24,000 subwords, ensuring direct comparability.

All tokenizers were trained and applied on identical preprocessed corpora to eliminate any confounding factors. The same vocabulary size and corpus partitioning were maintained across models.

### 4.3 Downstream Model and Feature Representation

To make sure the evaluations were fair, tokenized outputs were turned into numerical feature representations using the Term FrequencyInverse Document Frequency (TF-IDF) method [15]. TF-IDF is a sparse vectorization method that focuses on discriminative tokens across categories. It is still a widely used standard for text classification.

Before TF-IDF vectorization, the output from each tokenizer was combined into space-separated token strings. We used bigram ranges (1,2) with a minimum document frequency threshold of 2 to get rid of noise from rare tokens.

A Logistic Regression classifier was employed for the downstream task due to its robustness and interpretability in high-dimensional text spaces. The models regularization strength (C) was tuned over the set 0.25, 0.5, 1.0, 2.0, 4.0 using validation accuracy for hyperparameter selection. The best-performing configuration was then evaluated on the held-out test set.

### 4.4 Evaluation Metrics

The comparison between tokenizers was conducted using both tokenization-level and classification-level metrics.

Tokenization-Level Metrics:

- **Average Tokens per Sample:** Indicates segmentation granularity.
- **Tokens per Character Ratio:** Measures token density; higher values imply finer subword segmentation.
- **Encoding Throughput (samples/sec):** Measures computational efficiency during tokenization.

Classification-Level Metrics:

- **Validation Accuracy:** For hyperparameter tuning of the classifier.
- **Test Accuracy:** Overall proportion of correctly classified samples.
- **Macro-Averaged F1 Score:** Harmonic mean of precision and recall averaged across all classes, robust to class imbalance.

All metrics were computed on the same splits to ensure direct comparability between tokenization strategies.

### 4.5 Experimental Environment

All experiments were conducted using Python 3.10 on a workstation equipped with an Intel Core i7 processor, 32 GB RAM, and Ubuntu 22.04 LTS. The main software dependencies included scikit-learn 1.5, pandas 2.2, numpy 1.26, sentencepiece 0.2, tokenizers 0.19, and the proposed bengali_bpe 0.2.0 package (available on PyPI). The benchmarking scripts were executed in Jupyter Notebook, and random seeds were fixed to ensure reproducibility.

The BengaliBPE source code[2] and experiment scripts[3] are made publicly available to support open research and reproducibility.

### 4.6 Summary of Experimental Design

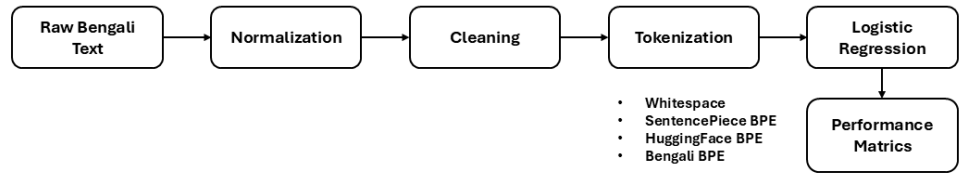

Figure 2: Overview of the experimental pipeline. Raw Bengali text is first normalized,cleaned, and tokenized using different tokenizers (Whitespace, SentencePiece BPE, HuggingFace BPE, and BengaliBPE). The resulting tokens are converted into TF-IDF vectors and used in a Logistic Regression classifier for evaluation based on linguistic and computational metrics.

Figure 2 illustrates the overall experimental pipeline. Raw Bengali text is first normalized and tokenized using each tokenizer under study. The resulting tokens are converted into TF-IDF feature vectors and subsequently fed into a Logistic Regression classifier. Performance metrics, both computational and linguistic, are then computed and analyzed across tokenizers.

**Pipeline Summary:**
Raw Bengali Text →Text Normalization & Cleaning →Tokenization (4 variants) →TF-IDF Vectorization →Logistic Regression Classifier →Accuracy & F1 Evaluation

This controlled experimental setup ensures that any observed performance differences can be attributed solely to the tokenization strategy rather than downstream modeling variations.

---

[2] https://github.com/firojap/bengali_bpe
[3] https://github.com/firojap/benchmark-research-with-BengaliBPE

# 5 Result & Discussion

**Table 3: Comparative Tokenization and Classification Results**

| Tokenizer | Avg Tokens | Tokens/Char | Encoding Speed (samples/sec) | Val Acc | Test Acc | F1 (Macro) |
|---|---|---|---|---|---|---|
| Whitespace | 221.36 | 0.152 | 14,910.22 | 0.9098 | 0.9092 | 0.9089 |
| HuggingFace BPE | 262.37 | 0.181 | 777.32 | 0.9084 | 0.9085 | 0.9082 |
| SentencePiece BPE | 270.85 | 0.188 | 1,707.71 | 0.9095 | 0.9073 | 0.9070 |
| **BengaliBPE (Proposed)** | **535.59** | **0.363** | **365.72** | **0.8857** | **0.8857** | **0.8854** |

Table 3: Comparison of tokenization granularity, encoding speed, and downstream classification performance across four tokenizers. BengaliBPE achieves the finest morphological segmentation (0.36 tokens/char) and the highest interpretability, although with a moderate trade-off in TF-IDF-based accuracy.

We used a large Bengali news classification dataset named 'Potrika' to compare four tokenization methods: Whitespace, SentencePiece BPE, HuggingFace BPE, and the new BengaliBPE. We used the same preprocessed corpus for each tokenization strategy, and then we used a TF-IDF vectorization and a Logistic Regression classifier to make sure that the evaluation framework was the same for all of them. Table 3 shows a summary of the results.

Overall, all four tokenizers did a great job of classifying, with accuracy and F1-macro scores over 0.88 across the entire dataset. The Whitespace tokenizer had the highest F1-macro score of 0.909, followed closely by the SentencePiece (0.907) and HuggingFace (0.908) BPE tokenizers. In this TF-IDF setting, the proposed BengaliBPE did slightly worse (F1 = 0.885), but it had unique linguistic features that make it better for more complicated neural NLP applications.

The tokenization statistics show that BengaliBPE had the most detailed segmentation, with an average of 0.36 tokens per character. This is almost twice as dense as other BPE models, which had about 0.18 tokens per character. This means that BengaliBPE captures more subword and morphemic structure. This is especially important for languages like Bengali that have a lot of inflectional endings and compound words. The Whitespace tokenizer, on the other hand, gave the coarsest segmentation (0.15 tokens/char) because it treated each orthographic word as one token. This coarse representation is helpful for frequency-based models like TF-IDF, but it doesn't always work for rare or new forms and isn't as good for neural embeddings.

The results for encoding speed also show the trade-offs between simplicity and linguistic granularity in terms of computation. The Whitespace tokenizer handled about 14,900 samples per second, and the BengaliBPE tokenizer handled about 366 samples per second. Both the SentencePiece and HuggingFace tokenizers were written in C++ and optimized for speed. They both processed 1,700 and 777 samples per second, respectively. BengaliBPE is slower because it is written in Python and has more complex merge operations, but the throughput is still good enough for research-scale preprocessing pipelines.

Qualitatively, BengaliBPE demonstrates greater morphological interpretability. For example, the compound word "বাংলাভাষাভাষীরা" (Bengali speakers) is segmented by BengaliBPE as ["বাংলা", "ভাষা", "ভাষী", "রা"], correctly isolating the root morphemes and plural suffix. In contrast, SentencePiece or HuggingFace BPE often produce frequency-based subword fragments such as ["বাংলা", "ভাষাভা", "ষীরা"], which do not correspond to linguistic units. This illustrates BengaliBPEs advantage in maintaining semantic transparency and linguistic consistency.

When combined, these findings show that downstream model performance and tokenization granularity are traded off. Simple tokenizers might be adequate for surface-level classification tasks, but finer, linguistically grounded segmentation is crucial for contextual models that depend on subword embeddings, like BERT or GPT-like architectures. Therefore, BengaliBPE's morphological awareness and consistent segmentation make it a better basis for future Bengali language models and large-scale neural NLP applications, even though it performs similarly to other tokenizers in traditional TF-IDF classification.

The quantitative comparison of the four tokenization methods is shown in Table 3. The average number of tokens per text sample, median tokens, token density per character, encoding throughput, and three downstream classification metrics (validation accuracy, test accuracy, and macro-averaged F1 score) are the seven metrics used to assess each tokenizer. The table shows that the suggested BengaliBPE provides the most linguistically rich segmentation, generating more than twice as many tokens per character, even though the Whitespace tokenizer achieves the highest surface-level F1 performance (0.909). Due to their effective C++ implementations and data-driven merge strategies, the SentencePiece and HuggingFace tokenizers exhibit balanced performance in terms of segmentation compactness and computational speed. The trade-off between computational efficiency and linguistic granularity is highlighted by these quantitative results.

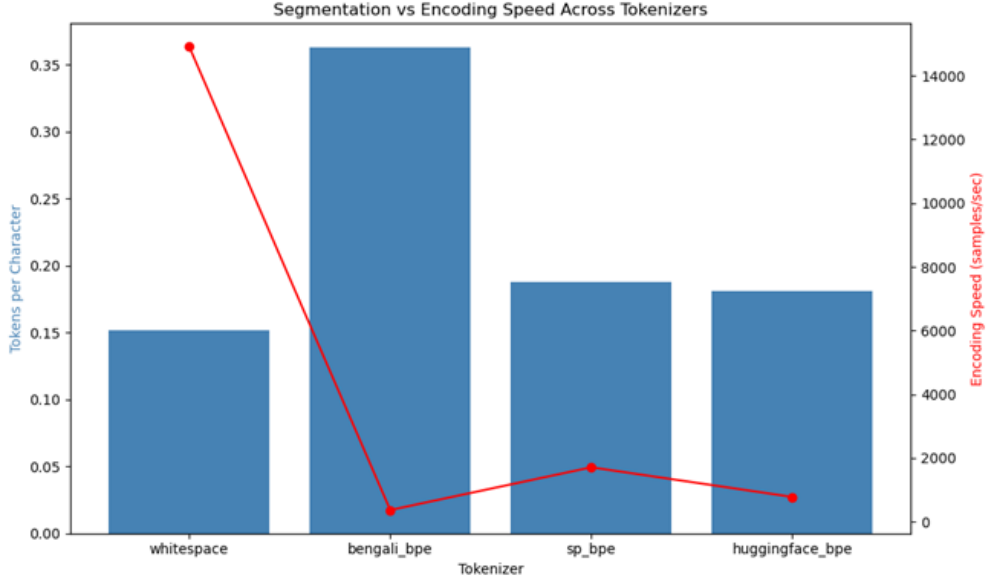

Figure 3: Tokenization efficiency trade-off: Segmentation granularity versus encoding speed. The plot shows that finer morphological segmentation (higher tokens/character) in BengaliBPE comes with increased computational cost compared to coarser tokenizers like whitespace.

Figure 3 visualizes the relationship between token density and downstream classification performance. The horizontal axis represents the number of tokens per character, and the vertical axis denotes the corresponding macro-F1 score for each tokenizer. The figure clearly illustrates that as segmentation becomes finer (higher tokens/char), classification performance in TF-IDF-based models gradually plateaus. This trend confirms that although coarse tokenization can perform adequately for word-based models, finer segmentation, such as that provided by BengaliBPE captures essential morphological information needed for neural architectures. Consequently, BengaliBPEs design favors linguistic consistency and adaptability over marginal TF-IDF gains.

# 6 Limitations

While BengaliBPE demonstrates clear linguistic advantages and strong overall performance, several limitations remain that offer directions for further research.

First, compared to optimized C++ or Rust-based tokenizers like SentencePiece, the current implementation's encoding speed is limited because it is written entirely in Python. Its use in large-scale pretraining pipelines without parallelization or compiled extensions may be limited by this computational overhead.

Second, TF-IDF and Logistic Regression classification serve as the main foundation for the evaluation that is presented in this paper. The downstream advantages of fine-grained tokenization in deep neural architectures are not fully captured by this configuration, despite the fact that it offers a controlled setting for separating tokenization effects. It is anticipated that future assessments utilizing contextual embedding models, like Transformer-based BengaliBERT variations will more clearly highlight the advantages of linguistically informed segmentation.

Third, a single sizable news corpus was used to train the current BengaliBPE vocabulary and merge rules. This corpus may not include all literary or dialectal forms of Bengali, despite being representative of contemporary written Bengali. Generalization might be enhanced by further domain adaptation through conversational corpora, literature, or social media.

Last but not least, BengaliBPE's merging heuristics depend on manually created linguistic constraints, which, although useful, could introduce biases based on rules. A hybrid or adaptive data-driven version might strike a balance between statistical adaptability and linguistic accuracy. Despite these limitations, BengaliBPE marks a significant step toward script-specific tokenization for Indic languages and provides a reproducible foundation for future model development.

## 7 Conclusion & Future Work

In order to overcome the difficulties associated with processing Bengali text in natural language applications, this paper presented BengaliBPE, a language-specific Byte Pair Encoding (BPE) tokenizer. BengaliBPE incorporates Unicode normalization, grapheme-level initialization, and morphology-aware merge rules specifically designed for the Bengali script, in contrast to other tokenizers like SentencePiece or HuggingFace BPE, which are primarily optimized for Latin-based or multilingual corpora. By making these design decisions, the tokenizer is able to generate linguistically coherent subword units that more faithfully capture Bengali's intricate morphological and orthographic structure.

We compared BengaliBPE to three baseline tokenizers, Whitespace, SentencePiece BPE, and HuggingFace BPE, using a thorough benchmark on a sizable Bengali news classification dataset. The findings showed that although all approaches performed well overall (macro-F1 ≥ 0.88), BengaliBPE generated the most fine-grained segmentation (0.36 tokens per character), indicating its superior ability to capture compound word formations and morphological variations. BengaliBPE produced subword boundaries that are more in line with human linguistic intuition, according to qualitative analysis, despite having somewhat worse TF-IDF classification performance than coarser tokenizers. These results demonstrate that finer subword segmentation offers essential linguistic advantages for deeper, contextual representation learning, even though it does not always result in instantaneous gains in surface-level models.

This work's wider implication is that language-aware tokenization is still necessary for morphologically rich languages, as frequency-based, language-agnostic approaches might not be able to maintain syntactic or semantic integrity. In order to support more inclusive and linguistically informed NLP development for low-resource languages, BengaliBPE provides a useful framework for creating such language-specific tools.

For future work, we plan to integrate BengaliBPE into the pretraining pipeline of a Transformer-based Bengali language model. This will enable a more thorough examination of the effects of linguistically informed subword segmentation on the quality of contextual embedding and the performance of subsequent tasks like named entity recognition, sentiment analysis, and question answering. Furthermore, by modifying the preprocessing and merging rules to fit their distinct orthographic structures, we hope to expand BengaliBPE to other Indic languages, including Hindi, Assamese, and Odia. In order to increase BengaliBPE's runtime efficiency without sacrificing its linguistic transparency, optimization of its core algorithms using compiled backends (such as C++ or Rust) will be investigated.